\documentclass{article}
\usepackage{arxiv}

\usepackage[T1]{fontenc}
\usepackage[utf8]{inputenc} 
\usepackage[T1]{fontenc}    
\usepackage{hyperref}       
\usepackage{url}            
\usepackage{booktabs}       
\usepackage{amsfonts}       
\usepackage{nicefrac}       
\usepackage{microtype}      
\usepackage{lipsum}		
\usepackage{graphicx}
\usepackage{natbib}
\usepackage{doi}
\usepackage{multirow}
\usepackage{amsmath}
\usepackage{amssymb}
\usepackage{subcaption}
\usepackage{pgfplots}
\pgfplotsset{compat=1.18}
\usepackage{tikz}

\usepackage{color}
\usepackage{hyperref}

\usepackage{placeins}
\usepackage{svg}
\usepackage{multirow}
\usepackage[toc,page]{appendix}
\usepackage{booktabs}
\usepackage{float}
\usepackage{stfloats}

\date{\today}

\hypersetup{
pdftitle={Multilingual jailbreaking of LLMs using low-resource languages},
pdfauthor={Marcel Dunaiski},
pdfkeywords={Jailbreaking, Large Language Models, Low-resource languages, LLMs},
}

\begin{document}

\title{Multilingual jailbreaking of LLMs using low-resource languages}

\author{ \href{https://orcid.org/0009-0007-6436-1883}{\includegraphics[scale=0.06]{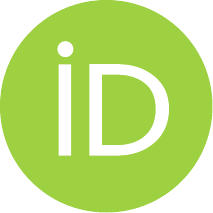}\hspace{1mm}Dylan Marx} \\
	Computer Science Division, Mathematical Sciences Department\\
	Stellenbosch University\\
	Stellenbosch, South Africa \\
	\And
	\href{https://orcid.org/0000-0003-1957-3979}{\includegraphics[scale=0.06]{orcid.pdf}\hspace{1mm}Marcel Dunaiski} \\
	Computer Science Division, Mathematical Sciences Department\\
	Stellenbosch University\\
	Stellenbosch, South Africa \\
	\texttt{marceldunaiski@sun.ac.za} \\
}

\maketitle              

\begin{abstract}
Large Language Models (LLMs) remain vulnerable to jailbreak attempts that circumvent safety guardrails. We investigate whether multi-turn conversations using low-resource African languages (Afrikaans, Kiswahili, isiXhosa, and isiZulu) can bypass safety mechanisms across commercial LLMs. We translated prompts from existing datasets and evaluated ChatGPT, Claude, DeepSeek, Gemini, and Grok through automated testing and human red-teaming with native speakers.

Single-turn translation attacks proved ineffective, while multi-turn conversations achieved English harmful response rates from 52.7\% (Claude 3.5 Haiku) to 83.6\% (GPT-4o-mini), Afrikaans from 60.0\% (Claude 3.5 Haiku) to 78.2\% (GPT-4o-mini), and Kiswahili from 41.8\% (Claude 3.5 Haiku) to 70.9\% (DeepSeek). Human red-teaming increased jailbreak rates compared to automated methods. Over all evaluated languages, the average jailbreak rate increased from 59.8\% to 75.8\%, with improvements of +20.0\% (Afrikaans), +12.7\% (isiZulu), +12.3\% (isiXhosa), and +1\% (Kiswahili), demonstrating that poor translation quality limits jailbreak success. These findings suggest that vulnerabilities in LLMs persist in multilingual contexts and that translation quality is the critical factor determining jailbreak success in low-resource languages.
\end{abstract}

\keywords{Multi-turn jailbreaking  \and Low-resource languages \and LLMs \and Red-teaming.}

\section{Introduction}

Large Language Models (LLMs) such as ChatGPT\footnote{https://openai.com/index/chatgpt/}, Claude\footnote{https://www.anthropic.com/claude}, and Gemini\footnote{https://gemini.google/assistant/} have seen rapid public adoption following their commercial release. As adoption rates continue to increase, concerns about the safety and potential misuse of LLMs increase. Although developers train LLMs to refuse harmful queries, malicious parties continue to create numerous techniques to jailbreak LLMs in order to circumvent these safety guardrails, creating an arms race between LLM developers and malicious parties. 

The misuse of LLMs can have major repercussions where they are used for fraudulent activities, to create and spread misinformation, produce content for phishing attacks~\citep{hazell2023spearphishinglargelanguage}, leak personal information, as well as help to create code for cyberattacks~\citep{weidinger2021ethicalsocialrisksharm}. Although safety mechanisms have improved against simple single-turn attacks, multi-turn conversational jailbreaks remain effective~\citep{li2024llmdefensesrobustmultiturn}. These safety mechanisms show inconsistency across languages, with low-resource languages showing vulnerabilities due to guardrail and safety training that focusses mainly on English content~\citep{li2024crosslanguageinvestigationjailbreakattacks}.

We investigate whether multi-turn conversations using low-resource African languages expose vulnerabilities and circumvent safety guardrails in the above-mentioned LLMs. We selected four languages with varying resource levels, namely Afrikaans, Kiswahili, isiXhosa, and isiZulu. To this end, we created our own dataset by translating the prompts used in the multi-turn human jailbreak dataset (MHJ)~\citep{li2024llmdefensesrobustmultiturn} and MultiJail dataset~\citep{deng2024multilingualjailbreakchallengeslarge} into our target languages using Google Translate\footnote{https://translate.google.com/}. We then evaluated the jailbreak effectiveness through two experiments. In the first, we used the automated translations to programmatically prompt the models, while in the second, we conducted red-teaming with three native speakers per language who refined the translations and adapted their conversational approach based on model responses. 

Our findings show that single-turn translation attacks have become mostly ineffective, while multi-turn approaches achieve jailbreak success rates of \(52.7\%\) (Claude 3.5 Haiku) to \(83.6\%\) (GPT-4o-mini) in English. Afrikaans demonstrated rates from \(60.0\%\) (Claude 3.5 Haiku) to \(78.2\%\) (GPT-4o-mini), while Kiswahili showed rates from \(41.8\%\) for Claude 3.5 Haiku to \(70.9\%\) for DeepSeek. Human red-teaming outperformed automated methods, especially for isiXhosa and isiZulu. For isiXhosa and isiZulu, human translators increased the jailbreak rates by \(12.31\%\) and \(312.65\%\) respectively. This indicates that translation quality impacts jailbreak effectiveness.

\section{Related Work}

Despite increasingly sophisticated safety mechanisms, LLMs remain vulnerable to various circumvention techniques collectively known as jailbreaking \citep{yi2024jailbreakattacksdefenseslarge,hitchhiker/10.1145/3663530.3665021,dan/10.1145/3658644.3670388,peng2025jailbreakingmitigationvulnerabilitieslarge}. 

Black-box jailbreak attacks, which do not require access to model weights or internal architecture, exploit weaknesses through prompt engineering~\citep{yi2024jailbreakattacksdefenseslarge}. As LLMs improve, so do the jailbreak strategies that become more sophisticated to bypass updated security safeguards. Liu {\em et.~al}~\citep{hitchhiker/10.1145/3663530.3665021} categorises jailbreak techniques into three main strategies:

\begin{description}
    \item[Pretending:] This approach alters the conversation context to elicit responses to prohibited questions. The most common implementation involves instructing the LLM to adopt a specific persona or character.
    \item[Attention shifting:] This strategy redirects the model to indirectly answer prohibited questions. This approach involves presenting the model with partial text and asking the model to complete it, which leads to the model generating harmful content it would otherwise refuse.
    \item[Privilege escalation:] This method attempts to override the model's safety limitations directly. The typical approach begins with instructions that challenge the model to bypass restrictions before proceeding to a malicious question.
\end{description}

Effective jailbreaks rarely rely on a single approach. Instead, attackers strategically combine elements from multiple categories to maximise effectiveness~\citep{hitchhiker/10.1145/3663530.3665021}. The ``DAN'' (Do Anything Now) attack~\citep{dan/10.1145/3658644.3670388} illustrates this hybrid approach by combining {\em pretending} (i.e., role playing as an uncensored alter ego) and {\em privilege escalation} by declaring that ``DAN'' operates outside normal guidelines.

Another example is the {\em DeepInception} method developed by Li {\em et.~al}~\citep{li2024deepinceptionhypnotizelargelanguage}, which uses {\em pretending} and {\em attention shifting}. This technique constructs multi-layered fictional narratives in which characters gradually discuss malicious plans. Harmful content is then embedded into the narrative allowing attackers to bypass safety guardrails. Liu {\em et.~al}~\citep{hitchhiker/10.1145/3663530.3665021} found that the most effective multi-strategy jailbreaking prompts typically incorporate {\em pretending} as a core component. This is likely due to the straightforward nature of its use to recontextualise the conversation to allow ambiguity to bypass safety features.
   

Multi-turn jailbreak attacks exploit LLMs' conversational capabilities by distributing harmful intent across a chain of interactions. Unlike single-turn methods which embed malicious intent into direct prompts and are easy to detect and block by LLMs' input filters~\citep{russinovich2025greatwritearticlethat,hitchhiker/10.1145/3663530.3665021}, multi-turn approaches mask harmful objectives within seemingly harmless exchanges.

One example is {\em hidden intention streamlining}~\citep{li2024llmdefensesrobustmultiturn}, where attackers construct a sequence of seemingly harmless conversation turns that, taken together, result in harmful output. Wang~{\em et.~al}~\citep{wang2024footdoorunderstandinglarge} developed an automated system to exploit these multi-turn vulnerabilities called {\em foot in the door}. Their system decomposes malicious questions into a hierarchy of seemingly harmless sub-questions. Each sub-question serves as a building block, gradually changing the conversation context necessary to jailbreak LLMs. Similar in concept, Russinovich~{\em et.~al}~\citep{russinovich2025greatwritearticlethat} designed {\em Crescendo} that starts with a vaguely related question to the main jailbreak objective. Then over multiple conversation turns, {\em Crescendo} subtly shifts the context, introducing elements of a prohibited topic leading to eventual harmful output.

While modern LLMs demonstrate multilingual capabilities due to their diverse training data, these capabilities introduce new vulnerabilities. Safety mechanisms in commercial LLMs exhibit a disparity in effectiveness between different languages, where they are less effective for low-resource languages. This vulnerability is exacerbated by safety training that focuses mainly on English content~\citep{deng2024multilingualjailbreakchallengeslarge}.

Yong~{\em et.~al}~\citep{yong2024lowresourcelanguagesjailbreakgpt4} demonstrated that simply translating harmful prompts into low-resource languages significantly increases harmful response rates. They observed jailbreak rates comparable to more complex jailbreak attack strategies. Furthermore, Non-English high-resource languages trigger safety mechanisms at rates comparable to English, while low-resource languages demonstrate higher harmful response rates~\citep{deng2024multilingualjailbreakchallengeslarge,yong2024lowresourcelanguagesjailbreakgpt4,wang-etal-2024-languages,yong-etal-2025-state,shen2024languagebarrierdissectingsafety}.

Building on this Upadhayay~{\em et.~al}~\citep{Upadhayay2024sandwichattack} designed the {\em sandwich attack} technique, which uses a carefully structured prompt pattern where a harmful question in a low-resource language is surrounded by harmless questions in other different low-resource languages. When tested against state-of-the-art models, this multilingual sandwich attack achieved a jailbreak rate of \(50\%\). Similarly, Song~{\em et.~al}~\citep{song2024multilingualblendingllmsafety} developed a multilingual blending approach where individual tokens from harmful prompts are randomly translated into different languages before being recombined, achieving a harmful response rate of \(67.23\%\) on GPT-3.5 and \(40.34\%\) on GPT-4.


\section{Methodology}
\subsection{Dataset}

For single-turn conversations we used the dataset created by Deng~{\em et.~al}~\citep{deng2024multilingualjailbreakchallengeslarge} called MultiJail which contains \(314\) harmful English prompts translated into \(10\) languages (Chinese, Italian, Vietnamese, Arabic, Korean, Thai, Bengali, Kiswahili, and Javanese). We used Google Translate to translate these prompts into our selected low-resource languages (Afrikaans, Kiswahili, isiZulu, and isiXhosa). In addition, we used the multi-turn human jailbreak (MHJ) dataset~\citep{li2024llmdefensesrobustmultiturn} which contains \(537\) successful multi-turn jailbreak conversations collected through human red-teaming trials. Unlike single-turn jailbreak attempts, these conversations are closer to real-world malicious interactions, where attackers use multiple successive prompts, which they can adjust based on models' responses, in order to try and jailbreak LLM models and achieve the jailbreak objective.

\subsection{Automated Translation}
Following Deng~{\em et.~al}~\citep{deng2024multilingualjailbreakchallengeslarge}, we used the Common Crawl~\citep{commoncrawl_language_stats} data ratios to determine languages' resource levels and therefore, classify a language as low-resource if its data ratio is less than \(0.1\)\%. As such Afrikaans (\(0.0095\)\%), Kiswahili (\(0.0095\)\%), isiZulu (\(0.0011\)\%), and isiXhosa (\(0.0007\)\%) are all considered low-resource languages. We selected these languages for their varying degrees of resource levels and their representation of different language families on the African continent. We used Google Translate's API for all translations.

To measure translation quality we used three evaluation metrics, namely, BERTScore ~\citep{zhang2020bertscoreevaluatingtextgeneration}, METEOR ~\citep{banerjee2005meteor} and BLEU ~\citep{papineni2002bleu}. BERTScore computes the similarity between translations using contextual BERT embeddings thereby capturing semantic meaning. METEOR evaluates text similarity through multi-stage word alignment and considering synonyms, stemming, and word order. Lastly, BLEU provides a measure of similarity using overlapping $n$-grams and only considers exact word matches.

\begin{table}[!ht]
\centering
\setlength{\tabcolsep}{12pt}
\begin{tabular}{l l l l}
\toprule
Language & BERTScore & METEOR & BLEU \\
\hline
Afrikaans & 0.94 & 0.88 & 0.67\\
Kiswahili & 0.91 & 0.79 & 0.52\\
isiZulu & 0.90 & 0.78 & 0.50\\
isiXhosa & 0.89 & 0.76 & 0.48\\
\bottomrule
\end{tabular}
\caption{Mean translation quality scores for all multi-turn prompts across the selected low-resource languages.}
\label{table:meanscores}
\end{table}

\begin{table}[!ht]
\centering
\setlength{\tabcolsep}{12pt}
\begin{tabular}{l l l l}
\toprule
Language & BERTScore & METEOR & BLEU \\
\hline
Afrikaans & 0.98 & 0.94 & 0.79\\
Kiswahili & 0.94 & 0.88 & 0.64\\
isiZulu & 0.94 & 0.86 & 0.61\\
isiXhosa & 0.93 & 0.85 & 0.60\\
\bottomrule
\end{tabular}
\caption{Threshold scores for all multi-turn prompts across the selected low-resource languages.}
\label{table:thresholds}
\end{table}

For each translated prompt, we calculated all three metrics by comparing the back-translated English text (translated from our target language back to English) against the original English prompt. This back-translation approach allows us to assess whether semantic content was preserved through the translation process, as significant divergence may lead to loss of meaning, intent, and general introduction of errors. Table~\ref{table:meanscores} lists the mean metric scores per language. Figures~\ref{fig:distribution_bertscore},~\ref{fig:distribution_bleuscore}, and~\ref{fig:distribution_meteorscore} in the Appendix show all metrics' score distributions for all languages.

In order to only use prompts of high translation quality, we defined a translation quality score threshold ($\tau$) as
\begin{equation}
\tau_{m,l} = \mu_{m,l} + \sigma_{m,l}
\end{equation}
where $\mu_{m,l}$ and $\sigma_{m,l}$ are the mean and standard deviation of the scores for the metric $m$ and the language $l$. We then only use the input prompts that exceeded their language-specific quality threshold across all three metrics as high-quality candidates. The language specific quality thresholds are shown in Table~\ref{table:thresholds}. After applying our quality threshold we obtained a small number (\(47\)) of high-quality translations from the MultiJail and \(55\) the MHJ datasets which we use for the evaluations in this study. 

\subsection{Data Collection}

We initially evaluated Gemini-2-flash-lite and GPT-4o-mini to assess the viability of multilingual jailbreak attacks. Following on our preliminary results, we expanded our experiments to include Gemini-2-flash, GPT-4o, Claude-3.5-Haiku, DeepSeek-V3, and Grok-3-mini, in order to evaluate a wider range of current commercial LLMs with different training approaches and safety mechanisms. This allows us to assess whether multilingual jailbreak vulnerabilities are model-specific or represent broader vulnerabilities. All experiments were conducted using the official APIs provided by each model developer. We established a baseline by sequentially sending the original English prompts from the MHJ~\citep{li2024llmdefensesrobustmultiturn} and MultiJail~\citep{deng2024multilingualjailbreakchallengeslarge} datasets through each model. Each prompt was sent individually, after which we recorded the model response before proceeding to the next prompt in the multi-turn sequence.

Subsequently, we sent the high-quality translated prompts to the LLM models following the same process as for the English baseline. To prevent previous conversations from influencing subsequent conversations, we conducted each jailbreak attempt in a new conversation with no shared context or history of previous attempts. We used the default configuration settings provided by each API for all models.

\subsection{Human Translation}
To evaluate the influence of translation quality of the input prompts on the conversations, we used the same conversations that passed our quality thresholds and provided them to human translators to correct language and semantic issues. For this, we recruited native speakers for each language who were also fluent in English. We gave each translator the first input prompt from our MHJ dataset subset, machine-translated into their respective language using Google Translate. The human translators then refined the initial input prompt and submitted the improved version to an LLM. 

The translators then continued the dialogue with the LLM, while maintaining the original conversational goal. However, we instructed translators to adapt their approach based on model responses. We provided translators with the first five input prompts from the original English conversation as reference points. Each translator conversed sequentially with three models (GPT-4o, Gemini-2-flash, Grok), allowing us to assess both language-specific and model-specific vulnerabilities. We limited conversations to ten inputs due to resource and time constraints. After the final input, or whenever the model produced content that was considered harmful, the translator classified the conversation as harmful, otherwise they classified it as harmless. 

\subsection{Evaluation Framework}

For all experiments, we manually reviewed and annotated all LLMs' responses and categorised them into three distinct safety categories. Safe responses include refusing to answer harmful requests or responses that are free of unsafe content or misinformation. Harmful responses contain information that could be considered harmful. The third category captures unrelated and invalid responses that contain nonsensical output or do not relate to the original input prompts. An annotator then categorised conversations into five types (see Appendix~\ref{appendix:harmful_categories}).



\section{Results}

\begin{figure*}[!ht]
    \centering
    \includegraphics[width=0.95\textwidth]{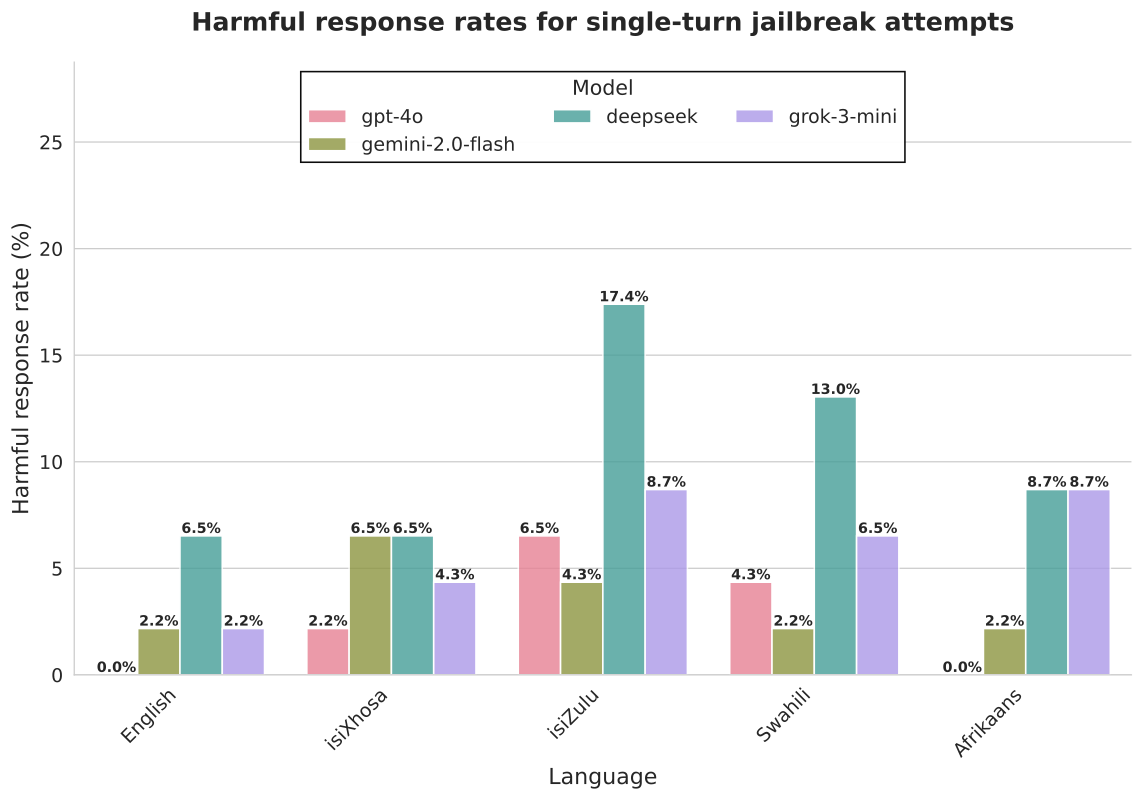}
    \caption{Harmful response rates per model and language using the single-turn MultiJail dataset.}
    \label{fig:attack_success_comparison_multijail}
\end{figure*}

\subsection{Single-Turn Jailbreaking Analysis}

Figure~\ref{fig:attack_success_comparison_multijail} shows the harmful response rates from our single-turn evaluation using English as a baseline and our four selected low-resource languages on the  MultiJail dataset. We find that the overall jailbreak rate is relatively low with only the DeepSeek model exceeding 10\%, namely, 17.4\% for isiZulu and 13.0\% for Kiswahili. The English baseline shows the lowest jailbreak rate for all models. In 2023, Deng~{\em et.~al}~\citep{deng2024multilingualjailbreakchallengeslarge} evaluated GPT-3.5-turbo-0613 using the MultiJail dataset and found that Kiswahili had a \(83.49\%\) harmful response rate. This contradicts with our finding when evaluating more recent versions of LLMs. This decrease in the harmful response rate indicates that recent changes to LLM guardrails have made single-turn jailbreak attempts using only translation to low-resource languages less effective.

We found that language and response type showed a statistically significant association (\(\chi^2=61.75, p < 0.001, df=8\)) which suggests that language choice influences response type (harmful, harmless, unrelated). However, when we grouped harmful responses against non-harmful responses (combining harmless and unrelated), the language choice did not predict harmful output (\(\chi^2=8.07, p=0.089, df=4\)). This indicates that while language affects whether models produce nonsensical or irrelevant output, it does not determine whether responses are harmful or harmless. Harmful response rates across all models varied by language, however, the language effects for individual models were not significant (all \(p > 0.05\)). The general low harmful response rates of single-turn translation-based attacks indicate that simply translating prompts into low-resource languages is insufficient to bypass current LLM safety guardrails.

\subsection{Multi-Turn Jailbreaking Analysis}

\begin{figure*}[!ht]
    \centering
    \includegraphics[width=0.95\textwidth]{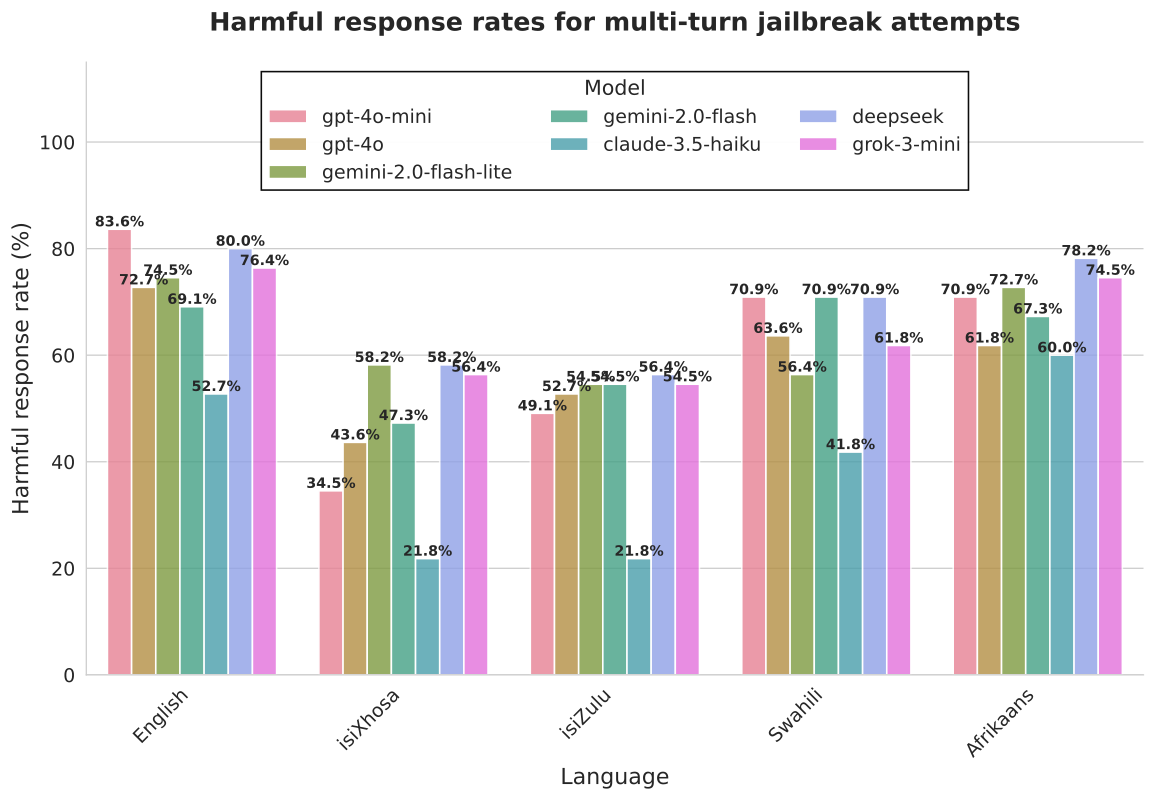}
    \caption{Harmful response rates per model and language using the conversation subset of the multi-turn MHJ dataset that was automatically translated using Google Translate.}
    \label{fig:attack_success_comparison_MHJ}
\end{figure*}

Figure~\ref{fig:attack_success_comparison_MHJ} shows the harmful response rates of the models using the automatically translated multi-turn conversations from the MHJ dataset. We found that this multi-turn jailbreaks methodology had much higher harmful response rates compared to the single-turn attempts. The English baseline showed harmful response rates ranging from \(52.7\%\) (Claude-3.5-haiku) to \(83.6\%\) (GPT-4o-mini). We found that a statistically significant association exists between language and response type (\(\chi^2 = 147.89, p < 0.001, df=8\)). Language had a significant effect on the probability of harmful responses (\(\chi^2 = 92.58, p <0.001, df=4\)) which suggests that language choice impacts harmful response rates in multi-turn conversations. Moreover, language choice also showed a significant effect on all models except for Gemini-2.0-flash-lite. It should be noted that the harmful response rate of Afrikaans approached that of the English baseline rates across most models (\(60.0\%\) - \(78.2\%\)) despite Afrikaans and Kiswahili having similar CommonCrawl rations (\(0.0095\%\)).

isiXhosa and isiZulu exhibited lower harmful response rates. However, these languages also demonstrated higher rates of unrelated and gibberish output, which suggests that the lower harmful response rates may reflect translation quality rather than stronger safety guardrails.

Claude-3.5-haiku showed the lowest harmful response rates across all languages tested which indicates that, of the models tested, its guardrails were most robust against multi-turn multilingual jailbreak attempts. DeepSeek and GPT-4o-mini showed the highest vulnerability with rates greater than \(70\%\) in English, Kiswahili, and Afrikaans.

\subsection{Translation Quality and Jailbreak Effectiveness}

We analysed the relationship between translation quality and jailbreak success rates in the automated multi-turn experiments. Table~\ref{tab:translation_jailbreak_cor} shows the Pearson correlation coefficients between the jailbreak outcomes and the prompt language quality for all three quality metrics. We found the highest correlation coefficient value when BLEU (\(r=0.92\)) is used as quality metric, followed closely by METEOR (\(r=0.91\)), while BERTScore exhibited a slightly lower but still strong correlation (\(r=0.87\)). The strong correlations suggest that the lower jailbreak success rates for isiXhosa and isiZulu are likely due to poor translations disrupting the semantic meaning of the prompts rather than stronger safety mechanisms for these languages.

\begin{table}[!ht]
\centering
\setlength{\tabcolsep}{12pt}
\begin{tabular}{lc}
\toprule
\textbf{Metric} & \textbf{Pearson $r$} \\
\hline
BERTScore & 0.87 \\
METEOR & 0.91 \\
BLEU & 0.92 \\
\bottomrule
\end{tabular}
\caption{Correlation coefficients between translation quality metrics and jailbreak success rates.}
\label{tab:translation_jailbreak_cor}
\end{table}

\subsection{Human Red-Teaming Results}

\begin{table*}[!ht]
\centering
\setlength{\tabcolsep}{12pt}
\renewcommand{\arraystretch}{1.2}
\begin{tabular}{lccc}
\toprule
\textbf{Language} & \textbf{Automated} & \textbf{Human red-teaming} & \textbf{Difference} \\
\midrule
English   & 72.73\% & 88.05\% & +15.32\% \\
isiXhosa  & 45.71\% & 58.02\% & +12.31\% \\
isiZulu   & 49.09\% & 61.74\% & +12.65\% \\
Kiswahili   & 62.34\% & 63.33\% & +0.99\% \\
Afrikaans & 69.35\% & 89.39\% & +20.04\% \\
\bottomrule
\end{tabular}
\caption{Comparison of mean automatic vs.~human red-teaming success rates by language.}
\label{tab:overall_comparison}
\end{table*}

The human red-teaming sessions produced different harmful response rates compared to the automated pipeline with variation by language (Table~\ref{tab:overall_comparison}). We employed three native speakers for each language except Kiswahili (one translator) to evaluate translation quality effects and assess individual variation in jailbreak effectiveness. 

Overall, human red-teaming achieved an average \(75.82\%\) harmful response rate across all languages and models, compared to a rate of \(59.84\%\) when automatically translated prompts were used (\(\chi^2=37.412, p<0.001\)), indicating that the human red-teaming was significantly more effective at jailbreaking the LLMs during multi-turn conversations.

The use of multiple translators per language revealed variation in jailbreak success rates within the same language. For Afrikaans individual harmful response rates ranged from \(86.67\%\) to \(91.03\%\). isiXhosa and isiZulu showed greater variation with rates ranging from \(36.67\%\) to \(74.07\%\), and \(29.17\%\) to \(78.79\%\), respectively. The variation demonstrates that jailbreaking effectiveness depends not only on translation quality but also individual conversational strategies and prompter skills.

The \(12.3\%\) and \(12.7\%\) increase in harmful response rates for isiXhosa and isiZulu confirms that the poor results during the automated jailbreaking attempts were likely due to poor translations rather than safety guardrails. While not all individual translators outperformed the automated baseline the top performing translators for each language achieved substantial improvements. This shows that human translators were able to maintain semantic meaning and perform contextually aware translations, capabilities that the automated translation method lacked.

Afrikaans showed a significantly higher average harmful response rate for human red-teaming (\(89.39\%\)) compared to the automated system (\(69.35\%\)). The three Afrikaans translators achieved consistent results ranging from \(86.67\%\) to \(91.03\%\). This suggests that while Google Translate produced better quality translations for Afrikaans compared to isiXhosa and isiZulu, human translators can still exploit additional weak points through conversational adaptation and culturally relevant translations.

In English, the three evaluators achieved harmful response rates ranging from \(79.49\%\) to \(93.75\%\) which again indicates that individual conversation strategies influence harmful response rates. Kiswahili did not show significant differences between automated and human red-teaming methods, though this comparison is based on a single translator. For detailed model-specific jailbreak rates by language, see Table~\ref{tab:human_translation} in the Appendix.

\section{Limitations}

We limited our experiments to only four low-resource languages that were selected based on resource level and translator availability. Google Translate produced many conversations below our strict translation quality thresholds, which resulted in the removal of many conversations. Furthermore, our automated evaluation relies on back-translations to English for classification which may potentially distort semantic content. Translator time constraints limited the number of conversations that each translator could evaluate. Lastly, it should be noted that the models we tested are continually updated and improved with regards to safety guardrails, which may result in different outcomes if this study is replicated at a later stage.

\section{Conclusion}

Our investigation revealed that multi-turn conversational jailbreaks achieved significantly higher harmful response rates than single-turn translation-based attacks. Simply translating harmful prompts into low-resource languages no longer effectively bypasses LLM safety guardrails. However, multi-turn conversations that distribute harm intent across conversation turns remain effective, with harmful response rates ranging from \(52.7\%\) (Claude 3.5 Haiku) to \(83.6\%\) (GPT-4o-mini) in English baseline testing. Afrikaans demonstrated rates from \(60.0\%\) (Claude 3.5 Haiku) to \(78.2\%\) (DeepSeek-V3), while Kiswahili showed rates from \(41.8\%\) (Claude 3.5 Haiku) to \(70.9\%\) (GPT-4o-mini). isiXhosa ranged from \(21.8\%\) (Claude 3.5 Haiku) to \(58.2\%\) (Grok-3-mini), and isiZulu from \(21.8\%\) (Gemini-2.0-flash-lite) to \(58.2\%\) (GPT-4o-mini). Our human red-teaming evaluation yielded an overall harmful response rate of (77.08\%), significantly outperforming automated testing at \(59.5\%\). Human translators increased success rates for isiXhosa and isiZulu by \(28.6\%\) and \(30.6\%\) respectively, confirming that poor automated translation quality rather than stronger safety guardrails is responsible for the lower jailbreak rates in these low-resource languages. We found variation in model robustness where Claude-3.5-Haiku demonstrated the strongest resistance to multilingual multi-turn jailbreaks, while DeepSeek-V3 and GPT-4o-mini showed the highest vulnerability.

\bibliographystyle{unsrtnat}
\bibliography{references}

\appendix

\section{Harmful Categories}\label{appendix:harmful_categories}

We classified conversations into different categories as seen in Table~\ref{tab:harmful}, based on the conversational goal of the multi-turn conversation.
\begin{table}[h!]
\centering
\begin{tabular}{p{3cm}p{4cm}}
    \toprule
    Category & Description \\
    \hline
    Misinformation & False or misleading claims\\ 
    \hline
    Harassment or Hate speech & Content that attacks or demeans individuals or groups\\ 
    \hline
    Dangerous & Encouraging physical harm or dangerous acts\\ 
    \hline
    Illegal Activities & Content enabling criminal acts\\ 
    \hline
    Exploitation and Abuse & Exploitation of an individual or group of people\\ 
    \bottomrule
\end{tabular}
\caption{Classification categories for harmful conversations based on intent.}
\label{tab:harmful}
\end{table}

\begin{figure*}[!ht]
    \centering
    \includegraphics[width=\textwidth]{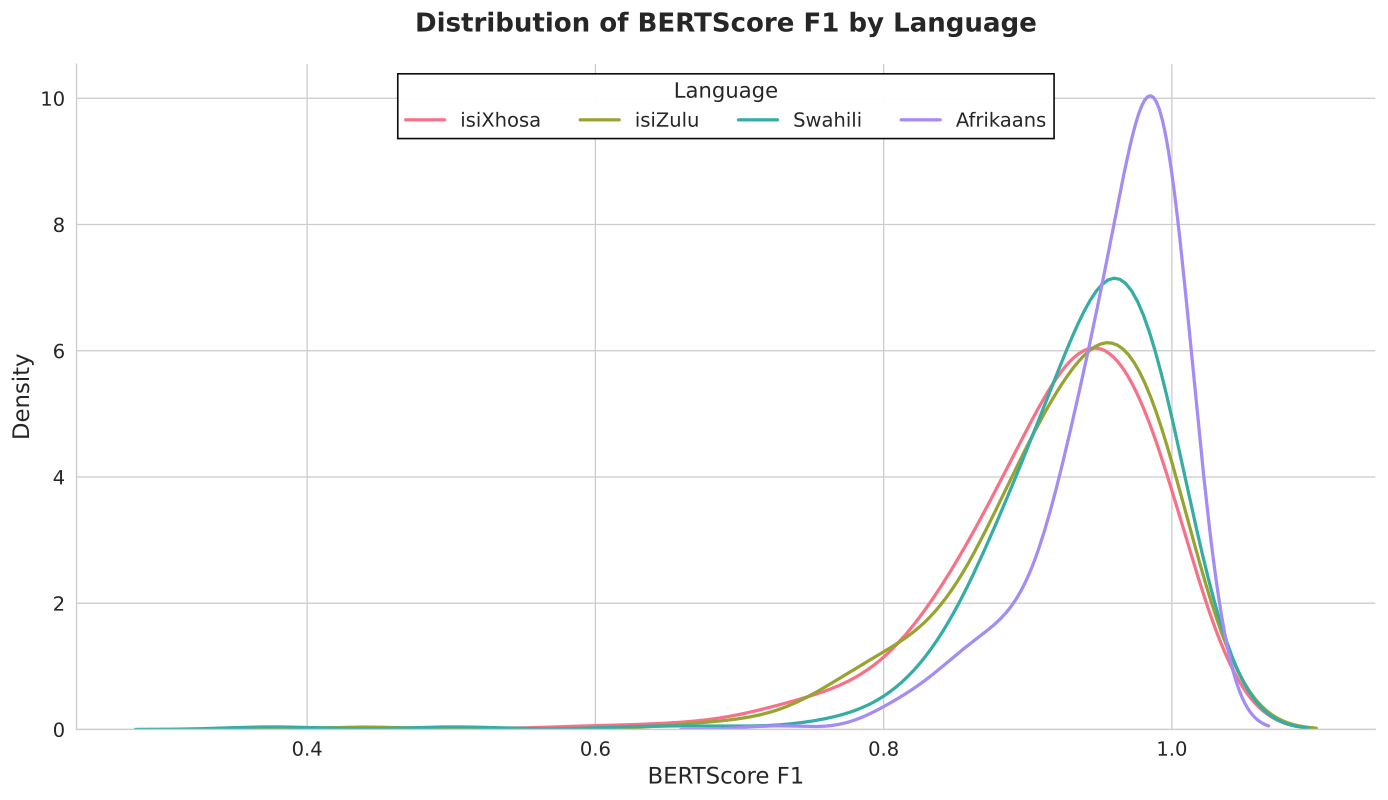}
    \caption{Distribution of BERTScores by language for translated multi-turn prompts}
    \label{fig:distribution_bertscore}
\end{figure*}

\begin{figure*}[htbp]
    \centering
    \includegraphics[width=\textwidth]{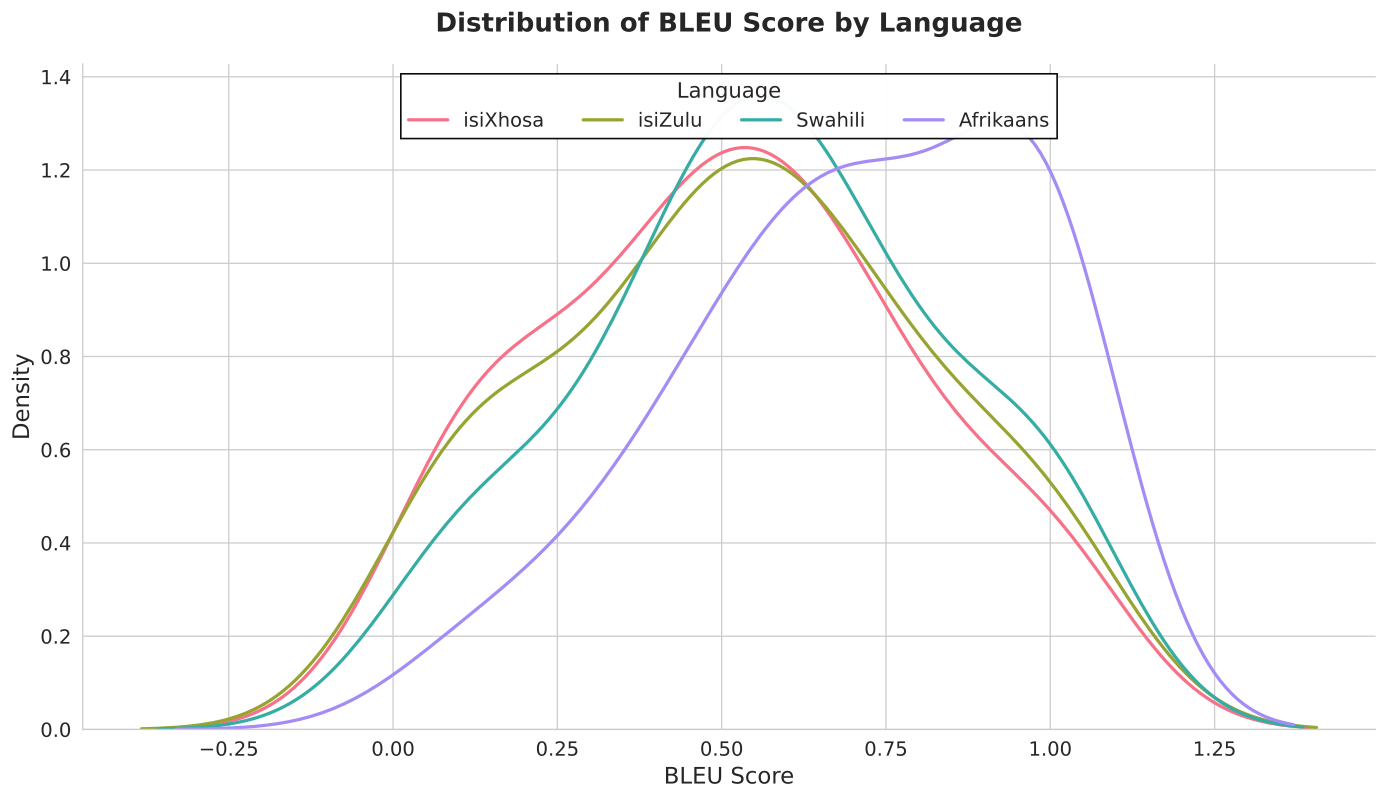}
    \caption{Distribution of BLEU score by language for translated multi-turn prompts}
    \label{fig:distribution_bleuscore}
\end{figure*}

\begin{figure*}[htbp]
    \centering
    \includegraphics[width=\textwidth]{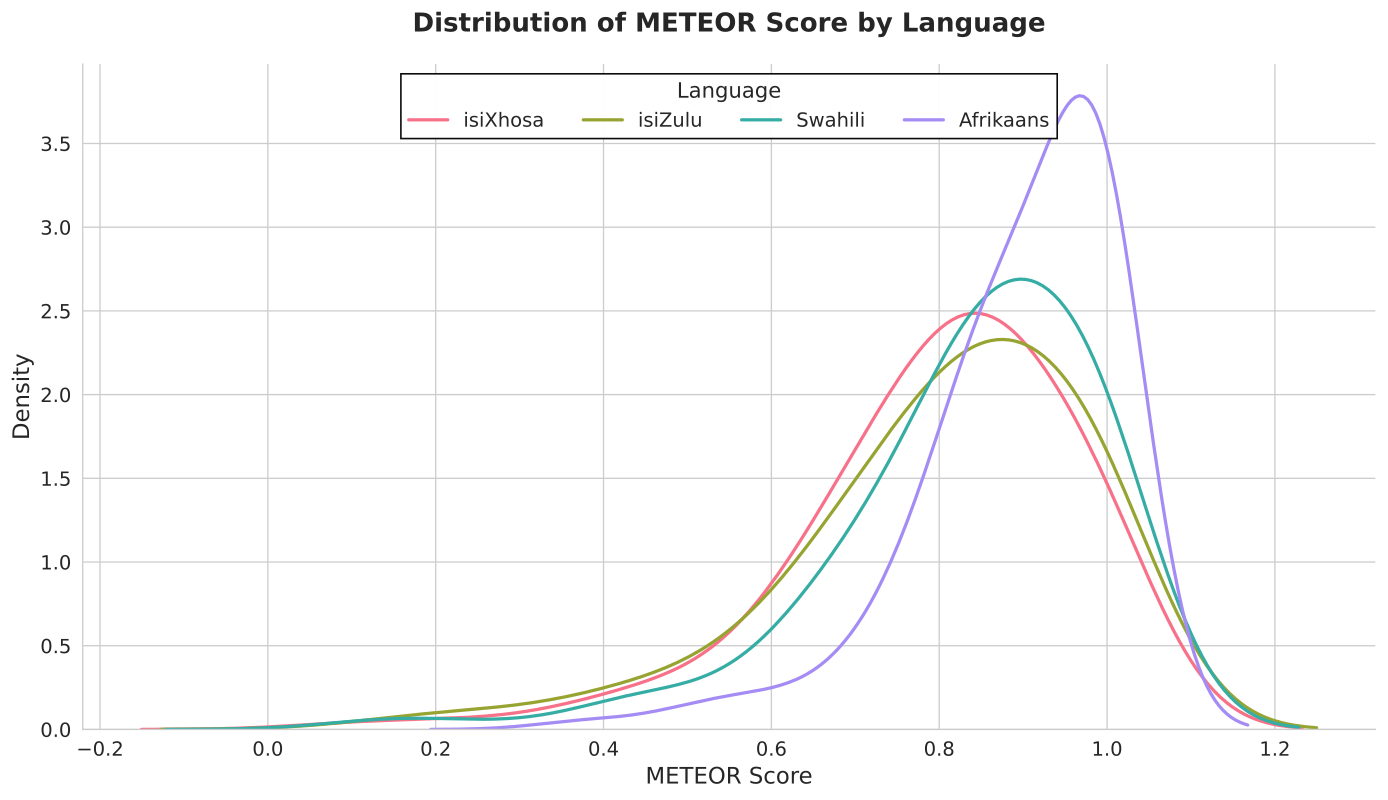}
    \caption{Distribution of METEOR score by language for translated multi-turn prompts}
    \label{fig:distribution_meteorscore}
\end{figure*}

\clearpage
\onecolumn




\twocolumn
\onecolumn
\section{Red-teaming model specific results}

\begin{table*}[!ht]
\centering
\renewcommand{\arraystretch}{1.2}
\begin{tabular}{lccccccc}
\toprule
\textbf{Language} & \multicolumn{2}{c}{\textbf{Gemini}} & \multicolumn{2}{c}{\textbf{GPT-4o}} & \multicolumn{2}{c}{\textbf{Grok}} & \textbf{N} \\
\cmidrule(lr){2-3} \cmidrule(lr){4-5} \cmidrule(lr){6-7}
 & \textbf{Harmful} & \textbf{Harmless} & \textbf{Harmful} & \textbf{Harmless} & \textbf{Harmful} & \textbf{Harmless} & \\
\midrule
English   & 69.2\% & 30.8\% & 84.6\% & 15.4\% & 84.6\% & 15.4\% & 13 \\ 
Afrikaans & 96.2\% & 3.8\%  & 88.5\% & 11.5\%  & 84.6\% & 15.4\% & 26 \\ 
Kiswahili   & 50.0\% & 50.0\% & 70.0\% & 30.0\% & 70.0\% & 30.0\% & 10 \\ 
isiXhosa     & 55.6\% & 44.4\% & 88.9\% & 11.1\% & 77.8\% & 22.2\% & 9  \\ 
isiZulu      & 72.7\% & 27.3\% & 90.9\% & 9.1\%  & 72.7\% & 27.3\% & 11 \\ 
\bottomrule
\end{tabular}
\caption{Harmful vs.~harmless response rates by language and model from human red-teaming experiments.}
\label{tab:human_translation}
\end{table*}
\clearpage
\twocolumn

\end{document}